\documentclass[10pt,twocolumn,letterpaper]{article}

\usepackage[pagenumbers]{wacv} % To force page numbers, e.g. for an arXiv version

\usepackage{graphicx}
\usepackage{amsmath}
\usepackage{amssymb}
\usepackage{booktabs}
\usepackage{makecell}
\usepackage[accsupp]{axessibility}  % Improves PDF readability for those with disabilities.

\usepackage[pagebackref,breaklinks,colorlinks]{hyperref}

\usepackage[capitalize]{cleveref}
\crefname{section}{Sec.}{Secs.}
\Crefname{section}{Section}{Sections}
\Crefname{table}{Table}{Tables}
\crefname{table}{Tab.}{Tabs.}

\def\wacvPaperID{1237} % *** Enter the WACV Paper ID here
\def\confName{WACV}
\def\confYear{2024}
\begin{document}

\title{Online Class-Incremental Learning For 
Real-World Food Image Classification}

\author{
Siddeshwar Raghavan\\
{\tt\small raghav12@purdue.edu}
\and
Jiangpeng He\thanks{corresponding author}\\
{\tt\small he416@purdue.edu}
\and
Fengqing Zhu \\
{\tt\small zhu0@purdue.edu}
\\ % This line break will separate the authors from the institution
School of Electrical and Computer Engineering, Purdue University, West Lafayette, Indiana USA
}

\maketitle

%%%%%%%%% ABSTRACT
\begin{abstract}

Food image classification is essential for monitoring health and tracking dietary in image-based dietary assessment methods. However, conventional systems often rely on static datasets with fixed classes and uniform distribution. In contrast, real-world food consumption patterns, shaped by cultural, economic, and personal influences, involve dynamic and evolving data. Thus, require the classification system to cope with continuously evolving data. Online Class Incremental Learning (OCIL) addresses the challenge of learning continuously from a single-pass data stream while adapting to the new knowledge and reducing catastrophic forgetting. Experience Replay (ER) based OCIL methods store a small portion of previous data and have shown encouraging performance. However, most existing OCIL works assume that the distribution of encountered data is perfectly balanced, which rarely happens in real-world scenarios. In this work, we explore OCIL for real-world food image classification by first introducing a probabilistic framework to simulate realistic food consumption scenarios. Subsequently, we present an attachable Dynamic Model Update (DMU) module designed for existing ER methods, which enables the selection of relevant images for model training, addressing challenges arising from data repetition and imbalanced sample occurrences inherent in realistic food consumption patterns within the OCIL framework. Our performance evaluation demonstrates significant enhancements compared to established ER methods, showing great potential for lifelong learning in real-world food image classification scenarios. The code of our method is publicly accessible at \href{https://gitlab.com/viper-purdue/OCIL-real-world-food-image-classification}{https://gitlab.com/viper-purdue/OCIL-real-world-food-image-classification}
\end{abstract}
\vspace{-0.4cm}
% %%%%%%%%%%%%%%%%%%%%%%%%%%%%%%%%%%%%%%%%%%%%%%% %%%%%%%%%%%%%%%%%%%%%%%%%%%%%%%%%%%%%%%%%%%%%%
\section{Introduction}
\label{sec:intro}

Food image classification has shown great potential for improving food pattern tracking~\cite{food_img_cls_deep_dnn, Liu2016DeepFoodDL, food_recog}, nutritional and health analysis~\cite{mao2021_nutri_hierarchy, NutriNet}, and dietary monitoring~\cite{yue_food_real_world, dietary-pattern-assess, auto_diet_assess, health-monitor}. Current food classification models have achieved remarkable performance on static datasets that do not change with time and have a fixed number of classes. However, food patterns change and food classes expand in the real world due to shifts in dietary styles and preferences~\cite{nutrition_dynamics, modern_trad_eating}. This makes it necessary for food image classification models to adapt and learn the new information from the changing data~\cite{he2023long}. In order to accommodate evolving data, we propose to leverage Online Class Incremental Learning (OCIL) approaches (outlined in detail in~\cite{ocil_survey} ). These methods continuously learn from a growing stream of sequentially arriving data. A major challenge in OCIL methods is catastrophic forgetting~\cite{catastrophic_forget_org, cat_forget, intransigence} of old knowledge while learning new data, leading to performance degradation. In this work, we focus on Online Class Incremental Learning (OCIL) frameworks~\cite{ocil_orig_1, CIL_online, food_ocil} for food image classification task. The OCIL setting is more realistic but challenging compared to offline, as the model must continually learn new classes from an online data stream encountering each incoming sample only once, while past observed data is not accessible. 

\begin{figure}[t]
  \centering
  \includegraphics[width=1.\linewidth]{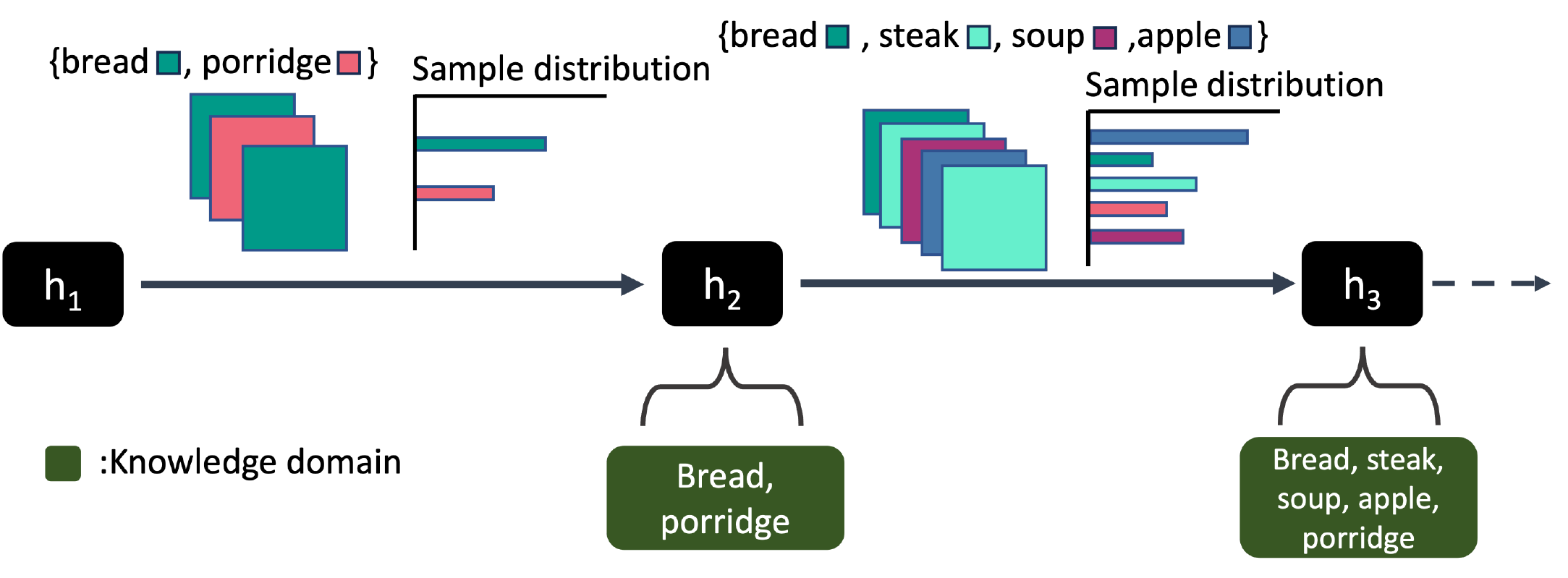}
  \caption{\textbf{R-OCIL for food image classification}. The model $h$ learns new food class data sequentially from realistic consumption patterns without accessing previously learned data. These data streams may include repetitions and imbalanced sample distributions. The updated model should classify all food classes it has encountered.}
  \label{fig:overall}
\end{figure}

Nevertheless, existing OCIL approaches operate under constraints that do not reflect real-world situations. For instance, certain constraints may impose equal task sizes, equal samples for each class in a task, and balanced training samples~\cite{MIR, GSS, DVC, ASER}. As a result, many existing methodologies assume that the training data is perfectly balanced, either explicitly or implicitly. However, real-world food consumption patterns seldom exhibit such fixed structure or balance. The most recent studies of Class-Incremental with Repetition (CIR) in offline scenarios \cite{cil_enough, hemati2023classincremental, pmlr-v119-chrysakis20a, gcil} have gained significant attention as a promising direction in Continual Learning. These settings offer enhanced flexibility in task definition, facilitating the integration of new and previously encountered classes. However, the consideration of repetition in this context remains under-explored in the online setting. This poses a challenge, but it holds significant relevance in practical, resource-constrained scenarios like the classification of food images in applications such as diet management and food intake monitoring systems \cite{yue_food_real_world, record-diet-assess}.

\begin{figure}
    \centering
    \includegraphics[width=1.\linewidth]{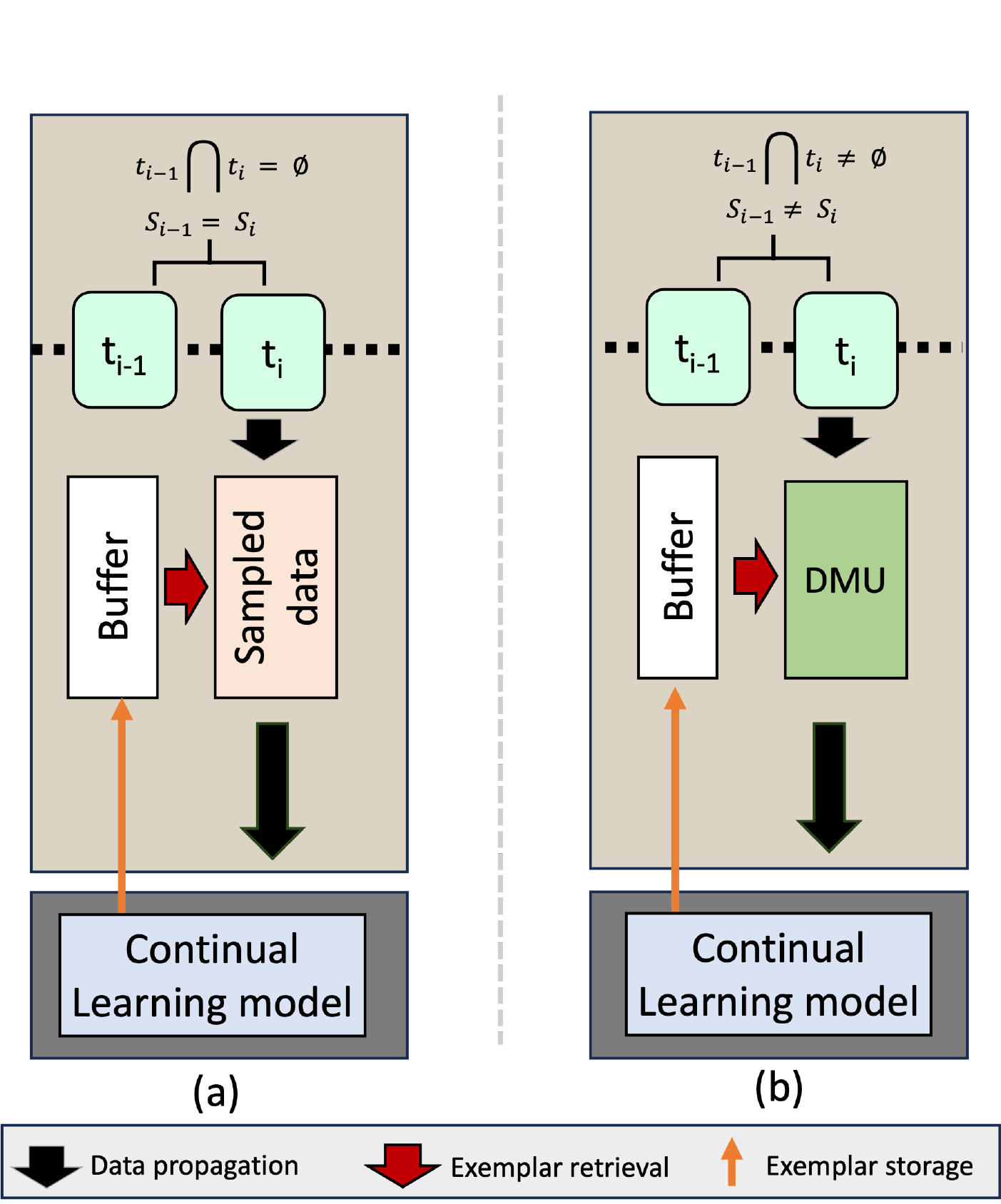}
    \caption{(a) Conventional ER OCIL strategy, (b) Our proposed realistic setting along with our DMU module. Existing OCIL ER studies~\cite{icarl,MIR,GSS,ASER,DVC,PCR} used (a). Each ER technique employs a different exemplar update and retrieval strategy, as indicated by the corresponding arrows in the figure.}
    \label{fig:origvsproposed}
\end{figure}

In this work, we fill this gap and extend the OCIL setting to a more realistic setup by removing the constraints mentioned above, which we refer to as Realistic Online Class Incremental Learning (R-OCIL) as shown in Figure \ref{fig:overall}. To generate experimental benchmarks that closely mimic real-world food patterns influenced by different dietary routines, we first summarize real-world food consumption patterns into three main categories, including short-term, moderate-term, and long-term. We introduce a Realistic Data Distribution Module (RDDM) to simulate these food consumption patterns. These scenarios are formulated based on approaches for safe and sustainable weight loss and maintenance, as explained in Section \ref{subsec:formulation}. We propose a simple yet effective plug-and-play module for existing Experience Replay (ER) OCIL methods called Dynamic Model Update (DMU) as shown in Figure~\ref{fig:origvsproposed}. The DMU optimizes the selection of the most representative samples for training, facilitating the natural accumulation of knowledge over time. 

The main contributions of our work can be summarized as follows:
\begin{enumerate}
    \item We propose a probabilistic framework called Realistic Data Distribution Module (RDDM) for Realistic Online Class Incremental Learning (R-OCIL) methods. To the best of our knowledge, this is the first such framework proposed for simulating various realistic food consumption patterns for image classification.
    \item We introduce a dynamic plug-and-play module called Dynamic Model Update (DMU), which can be integrated with existing Experience Replay (ER) OCIL methods to improve learning accuracy and mitigate catastrophic forgetting in realistic scenarios.
    \item We simulate three real-world food consumption patterns - short-term, moderate-term, and long-term on the challenging Food-101~\cite{food101} and VFN~\cite{he2022longtailed} datasets using our proposed RDDM framework. We show that incorporating our DMU module significantly improves the performance of ER methods across all three scenarios.
\end{enumerate}
% %%%%%%%%%%%%%%%%%%%%%%%%%%%%%%%%%%%%%%%%%%%%%%% %%%%%%%%%%%%%%%%%%%%%%%%%%%%%%%%%%%%%%%%%%%%%%
\section{Related Work}
\label{sec:related_work}

\subsection{Food Image Classification}
Food image classification is a computer vision task that assigns a food category or label to an input food image. This task falls under the umbrella of image classification, wherein deep learning models are trained to recognize different food items based on their visual features as in~\cite{NutriNet, auto_diet_assess, dl_food_class2, dl_food_class3, foodai_logging, Liu2016DeepFoodDL, food_img_cls_deep_dnn}. The task typically involves assigning a single food category to an input image, assuming that the image contains only one type of food. Furthermore, recent research efforts have leveraged hierarchical structures~\cite{dl_food_class4} based on visual information to achieve further performance improvements. Nonetheless, all these methodologies rely on fixed food image datasets for training, lacking the capacity to learn from sequentially presented data. This limitation hampers their applicability in real-world scenarios where new foods are continuously encountered over time.

\subsection{Online Class Incremental Learning}

Class Incremental Learning (CIL) ~\cite{cil1, cil2, van_types_of_CL, Belouadah2019IL2MCI} is a machine learning paradigm where a model is trained to progressively incorporate new classes or categories of data over time. This learning approach involves updating the model in such a way that it can accommodate additional classes without compromising its ability to recognize previously learned classes. In the context of CIL, there are two primary learning protocols: (i)\textit{Offline}~\cite{CIL_survey, cil_enough, hemati2023classincremental} (We refer to as CIL in this paper)(ii) \textit{Online}~\cite{ocil_survey, CIL_online}(We refer to as OCIL in this paper): In this protocol, the training data also arrives sequentially, similar to the offline scenario, but the model can only be trained on each sample once without accessing data from the previous tasks.
Current OCIL methods can be taxonomized into two major categories: (i)Regularization-based~\cite{synaptic_CL, better_weight_cons, EWC, AGEM, GEM, LWF} and (ii) Experience Replay-based methods. Regularization-based techniques involve including an extra penalty term within the loss function to penalize updates to important parameters. 
Experience Replay-based (ER) techniques address the issue of catastrophic forgetting by storing a subset of learned task data as exemplars within a memory buffer. These exemplars are then employed for rehearsal during the process of continual learning. When compared to regularization-based methods, ER techniques have showcased superior efficiency. In this study, we center our attention on ER-based approaches for comparison. The MIR~\cite{MIR} method executes a virtual parameter update and retrieves data from the memory buffer that are primarily impacted, using the loss from the current mini-batch. Additionally, the buffer retrieval and storage procedures vary significantly between the``online" and ``offline" settings due to the specific data access constraints.

In contrast, GSS~\cite{GSS} opts for exemplar selection based on gradient directions. Although initially designed for CIL(offline), iCaRL~\cite{icarl, ocil_survey} has exhibited effectiveness even in the OCIL(online) context, achieved through random exemplar selection coupled with the nearest-class-mean classifier. ASER~\cite{ASER}, singles out the most efficient data for use as exemplars using adversarial shapley value. DVC~\cite{DVC} introduces a gradient-based exemplar selection process that primarily responds to incoming samples with high interference. This approach also includes a representation learner that maximizes mutual information, drawing inspiration from contrastive learning. 

% %%%%%%%%%%%%%%%%%%%%%%%%%%%%%%%%%%%%%%%%%%%%%%% %%%%%%%%%%%%%%%%%%%%%%%%%%%%%%%%%%%%%%%%%%%%%%
\section{Problem Formulation}
\subsection{Problem setup}
Online Class Incremental Learning involves learning a sequence of $T$ tasks, denoted by $t_1$ to $t_T$, one task at a time. During the learning of any task $t_i$, no data from any other task other than $t_i$ is accessed. The model is updated from $h_1$ to $h_T$. In each task $t_i$, there are $K_i$ classes, and the distribution of samples and classes adheres to the three characteristics detailed in Section \ref{subsec:chars_R-OCIL}. 
For a specific task $t_i$, the training data for the model is denoted as $S_i$, which consists of $n_i$ sets of training examples presented in the structure $(x_{i,1}, y_{i,1}),...(x_{i, {n_i}}, y_{i, {n_i}})$. Here, $x$ refers to an image, and $y$ represents its associated label. The index $i$ corresponds to the particular task $t_i$, where $i \in \{1, 2,...,T\}$.

\subsection{Characteristics of R-OCIL}
\label{subsec:chars_R-OCIL}
We identify three main characteristics of R-OCIL and use them to propose a probabilistic formulation to simulate realistic food consumption patterns.
Let $C = \{C_1, C_2,...,C_N\}$ denote a set of $N$ classes. The sample sizes of different classes appear in a task $t_i \in T$ are modeled as a random vector $S_{t_i}$ where each element $s_{i, j}$ represents the sample size of class $C_j$ in task $t_i$. We normalize the total samples of a class $C_j$ by its occurrences across all tasks $T$.

\noindent\textbf {Characteristic 1: } \label{char1}The total number of classes in a given task $t_i$ is not fixed. We denote the number of classes within that task as $K_i$, expressed as:
\begin{equation}
    K_{i} = i \in C: 1 \leq \sum_{j}^{} s_{i,j} < N 
    \label{eqn:char_1}
\end{equation}
To uphold the concept of class incremental learning, we establish $K_i \geq K_{i-1}+1$, ensuring that each task $t_i$ introduces at least one new class compared to the preceding task $t_{i-1}$. Additionally, by setting an upper limit below $N$ (the overall number of classes), we prevent all food classes from being presented within a single task. This scenario, uncommon in reality, would lead to the transformation of the problem into Domain Incremental Learning \cite{van_types_of_CL}. In such cases, the model would need to process all data simultaneously.

\noindent\textbf{Characteristic 2: }\label{char2} The sample sizes of occurring classes within any given task $t_i$ may not be equal.
\begin{equation} 
\begin{gathered}
 \forall m, n \in N \mid m \neq n \\
 P(s_{i,m} \neq s_{i,n}) > 0, \; \forall \; s_{i,m} \neq s_{i,n} \neq 0
\label{eqn:char_2}
\end{gathered}
\end{equation}

\noindent\textbf{Characteristic 3: }\label{char3} Classes appearing across different tasks could overlap. Let's consider two tasks $t_i, t_j$ such that $i \neq j $

\begin{equation}
    C_{t_i} \cap C_{t_j} \neq \emptyset
    \label{eqn:char_3}
\end{equation}
where $C_{t_i}$ and $C_{t_j}$ denote the classes occurring in tasks $i$ and $j$.

\subsection{Formulation of food consumption pattern categories}
\label{subsec:formulation}
Food consumption patterns offer significant insights into the dietary behaviors of individuals or communities within a defined time frame. This encompasses details like the kinds and quantities of ingested food and beverages, eating routines, meal frequencies, and food preparation methods. Understanding these patterns is essential for delving into the relationship between diet and health consequences \cite{auto_diet_assess, NutriNet, dietary-pattern-assess}. 

Due to the unavailability of datasets capturing food consumption patterns, we simulate them. Nonetheless, considering the multitude of potential permutations in food consumption patterns, replicating every single one proves to be a formidable task. To streamline this process, we classify these patterns into three primary types: \emph{short-term}, \emph{moderate-term}, and \emph{long-term}. These classifications draw from dietary and nutritional recommendations~\cite{diet_approach_durations, diet_1}.

In reality, food consumption patterns often do not involve the consumption of every available food item. To address this, we introduce two hyperparameters labeled $\alpha$ and $\beta$. The hyperparameter $\alpha$ controls the repetition of previously observed classes and $\beta$ controls the addition of new food classes. We must consider diverse dietary styles to determine suitable values for these hyperparameters corresponding to each food consumption category. For instance, certain diets like low-protein regimens, which restrict caloric intake, are typically followed for short duration, around six months, to facilitate healthy weight loss without negative consequences~\cite{diet_approach_durations}. To simulate such a diet within the short-term food consumption category, we cap the number of accessible food classes at 40 for the Food-101 dataset and 30 for the VFN dataset, respectively.
Thus, although $\alpha$ and $\beta$ offer a broad spectrum of feasible values for encompassing various food consumption patterns, we have selected three sets of values as representatives of these dietary habits. These specific values are elaborated upon in Table \ref{table:category}.

\begin{table}
\caption{\textbf{Food consumption categories}. The range of classes for each dataset experimented with and the corresponding $\alpha$ and $\beta$ values used to produce patterns reflecting the number of represented classes.}
\label{table:category}
\resizebox{\columnwidth}{!}{
\begin{tabular}{@{}lcccc@{}}
\toprule
\begin{tabular}[c]{@{}c@{}}Food \\ Consumption \\ Category\end{tabular} & $\alpha$               & $\beta$          & \begin{tabular}[c]{@{}c@{}}Food 101\\ 101 Classes\end{tabular} & \begin{tabular}[c]{@{}c@{}}VFN\\ 74 Classes\end{tabular} \\ \midrule
Short-term                                                              & {[}9, 10, 11, 12{]} & {[}1, 2, 3{]} & 20-40                                                          & 10-30                                                    \\
Moderate-term                                                           & {[}5, 6, 7, 8{]}    & {[}4, 5, 6{]} & 40-80                                                          & 30-50                                                    \\
Long-term                                                               & {[}1, 2, 3, 4{]}    & {[}7, 8, 9{]} & 80-101                                                         & 50-74                                                    \\ \bottomrule
\end{tabular}
}
\end{table}

% %%%%%%%%%%%%%%%%%%%%%%%%%%%%%%%%%%%%%%%%%%%%%%% %%%%%%%%%%%%%%%%%%%%%%%%%%%%%%%%%%%%%%%%%%%%%%
\section{Methodology}
\label{sec:method}

%%%%%%%%%%DMU+main pipeline figure%%%%%%%%%%%%%%%%

\begin{figure*}[t]
    \centering
    \includegraphics[width=\textwidth]{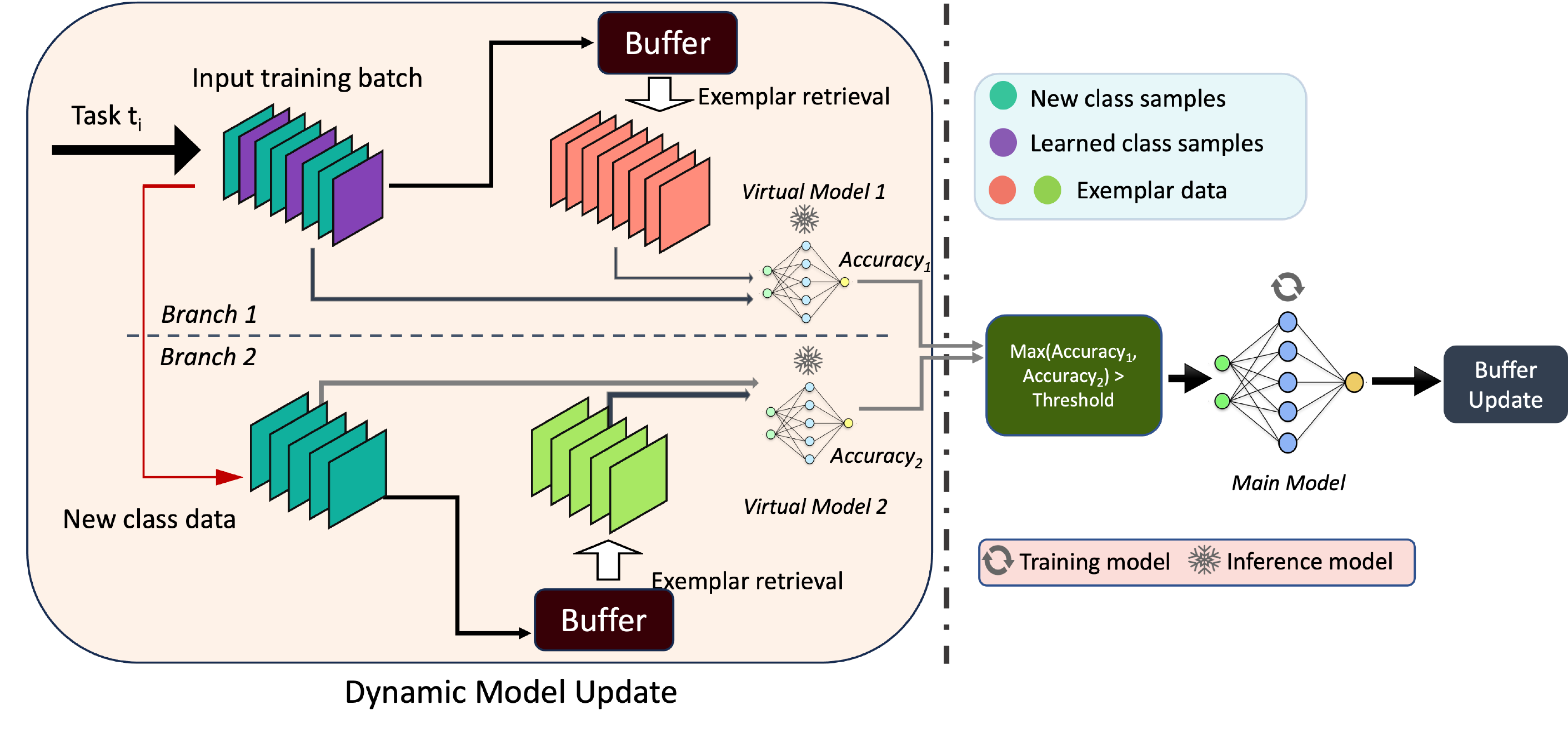}
    \caption{\textbf{Overview of proposed method}: The left section of the diagram illustrates the Dynamic Model Update (DMU) process, containing two concurrent data inference pathways. The first branch encompasses a virtual model denoted as $VM_1$, which receives the complete batch of data and the corresponding exemplars retrieved from the buffer. Simultaneously, the second branch involves another virtual model named $VM_2$, and receives only the non-repeated data (associated with new class samples) computed based on Equations~\ref{eqn:overview} and \ref{eq:non_overlapped_samples} along with its corresponding exemplars. At any given task $t_i$, the virtual models correspond to the OCIL model trained until task $t_{i-1}$. The strategy for retrieving exemplars and updating the buffer depends on the specific ER method employed in the experiment. The accuracy of these two virtual models is subsequently compared against the accuracy of the initial task, serving as the threshold. Next, the input image sequence and exemplars from the virtual branch that exhibits superior performance are directed into the main OCIL model for training.}
    \label{fig:pipeline}
\end{figure*}

Our method comprises two primary components. Firstly, we introduce a probabilistic framework called the Realistic Data Distribution Module (RDDM). This module aims to simulate realistic food consumption patterns, as elaborated in Section~\ref{subsec:formulation}, while aligning with the characteristics of Realistic OCIL as outlined in Section~\ref{subsec:chars_R-OCIL}. Secondly, we propose a plug-and-play module named the Dynamic Model Update (DMU). This module is designed to seamlessly integrate with existing experience replay-based OCIL methods. The DMU operates by continuously assessing the model's current performance and dynamically selecting the most representative food images from the input image sequence within each task for training. This adaptive approach enhances overall learning and reduces forgetting.

\subsection{Realistic Data Distribution Module (RDDM)}
\label{subsec:rddm}
Simulating food consumption patterns can be challenging due to managing multiple variables and determining suitable repetition and probabilistic distributions. Additionally, it can be difficult to select which food classes to repeat, their order and distributions. In this paper, we introduce a framework for generating realistic food scenarios that can be customized to various dietary patterns by changing the input probability distribution and two hyperparameters. The proposed framework aims to simplify the process of generating realistic food consumption patterns, which can be helpful in various applications, such as food recommendation systems, continual health monitoring, and nutrition research.

We consider an RDDM formulation for realistic food consumption patterns following the three characteristics introduced in Section~\ref{subsec:chars_R-OCIL}.
\begin{equation}
    \begin{gathered}
        K_{i}^{repeat} = D(i, \alpha) \\
        K_{i}^{new} = D(i, \beta) \\
        K_{i} = K_{i}^{new} +K_{i}^{repeat}
    \end{gathered}    
    \label{eqn:class_num}
\end{equation}

The task-specific distribution $D$ determines the \textbf{number of classes} $K_{i}$ present in a given task $i$. We split this equation into two parts. As described earlier, we are using the hyperparameter $\beta$ to control the addition of new food classes and the hyperparameter $\alpha$ to control the repetition of previously observed classes. As a result, $K_{i}$ is a stochastic variable that fulfills \hyperref[char1]{\textbf{Characteristic 1}}. Different food consumption patterns can be simulated by selecting various data distributions $D$. For instance, in Section \ref{subsec:ablation}, we employ an exponential function as the data distribution $D$ to generate a more significant level of imbalance. The degree of imbalance, represented by the imbalance factor ($\rho$), relies on the number of categories present in the dataset and the hyperparameters $\alpha$ and $\beta$ as represented in Equation~\ref{eqn:imbalance}.
\vspace{-0.1cm}
\begin{equation}
    \rho = \frac{N - E}{\alpha}
    \label{eqn:imbalance}
\end{equation}  
where $N$ represents total number of classes and $E$ represent the number of encountered classes. 
\begin{equation}
    A_{i} = L(K_{i}, W_{i}^{'})
    \label{eqn:appear_class}
\end{equation}
The binary indicator vector $A_{i}$ denotes which classes appear in task $t_i$ and is drawn from distribution $L(K_{i}, W_{i}^{'})$ shown in Equation~\ref{eqn:appear_class}, where the value "1" indicates the presence of a particular class. This distribution is dependent on the number of classes $K_{i}$ and satisfies \hyperref[char3]{\textbf{Characteristic 3}}. The weight vector $W_{i}^{'}$ determines the probability of a class appearing in task $t_i$, with higher weights indicating a greater likelihood of appearance. We set $L$ as uniform distribution in Section~\ref{subsec:exp_setup} as each class has an equally likely chance of appearing in a task.
\vspace{-0.2cm}
\begin{equation}
        S_{i} = Q(A_{i})
     \label{eqn:sample_size}   
\end{equation}
$S_{i}$ represents the \textbf{sample sizes} of each class in $A_{i}$, and is drawn from the distribution $Q(A_{i})$ as shown in Equation~\ref{eqn:sample_size}. To simplify the setup, we allocate the total number of samples for a class by dividing it by the total number of occurrences across all tasks and satisfies \hyperref[char2]{\textbf{Characteristic 2}}. For example, if \textit{``class 5"} has 500 samples and appears five times in ten tasks, each appearance of \textit{``class 5"} in a task will have 100 samples. Any extra samples are allocated to the first task if this number is not divisible.

\subsection{Dynamic Model Update (DMU)}
\label{subsec:DMU}
In this section, we introduce the Dynamic Model Update (DMU), as depicted in Figure~\ref{fig:pipeline}. Given the common occurrence of repeated foods in real-life food consumption patterns, which can act as natural replays and partially alleviate the impact of forgetting, we designed a straightforward and effective training module. The integration of DMU with any existing ER method can be seamlessly achieved without disrupting its workflow. Our primary objective is to enhance the model update pipeline to integrate seamlessly with the latest ER methods. Additionally, we showcase how our approach can be adapted to accommodate diverse buffer update strategies employed by various ER methods, resulting in enhanced performance for real-world food classification.

The key idea of our DMU is to avoid training the model without being aware of the input image data from the sequence. Moreover, since the input sequence can contain possible repetitions of previously encountered classes, the model could overfit if images from a specific class appear frequently. To address this, we have devised a strategy that dynamically selects whether to incorporate the entire training batch or solely the new classes based on the model's ongoing training performance. Upon encountering data from task $t_i$ during the training phase, we establish two virtual models: $VM_1$ and $VM_2$. All layers are frozen in these virtual models, and gradient updates are not applied. These virtual models are clones of the primary model trained until the current task $t$, denoted as $h_{i-1}$. The first branch ($VM_1$) of these models is deduced using the complete input batch $[X, Y]$ and corresponding exemplars $[X_E, Y_E]$, as depicted in Equation \ref{eq:vm1}.

For $VM_2$, the second branch of the models, the input comprises $[X_H, Y_H]$, alongside corresponding exemplars retrieved from the buffer, represented as $X_{E_H}, Y_{E_H}$. The selection of exemplars for the two virtual branches differs due to the distinct inputs of each branch. Each ER method employs a distinct exemplar retrieval strategy, ranging from random retrieval in iCaRL and GSS to constraint-based retrieval in MIR, ASER, and DVC. To acquire exemplars for the non-repeated (new sample) class images, the prior label distribution ($PD(t_i)$) and the current input $[X, Y]$ are compared. Here, $PD(t_i)$ denotes the stored values of all unique label sets that have emerged until task $t_i$. The computation of non-repeated data $[X_H, Y_H]$ can be accomplished using Equation \ref{eqn:overview}, as illustrated in Figure \ref{fig:pipeline}.

\begin{equation}
    \begin{gathered}
        \hat{y}^{VM_1} = VM_1^{h_{i-1}} ([X,  X_E]) \\
        prediction(x_i) = argmax(\hat{y}^{VM_1}_i) \\
        Accuracy^{VM_1} = \frac{1}{N_1}\sum_{i=1}^{N_1}[\hat{y}^{VM_1} = {y}^{VM_1}]
        \label{eq:vm1}
    \end{gathered}  
\end{equation}
Where $\hat{y}^{VM_1} = [Y,  Y_E]$ and  $N_1$ is the  corresponding batch size.
\vspace{-0.1cm}
\begin{equation}
    H([X,Y], PD(k)) = [X_H, Y_H]
    \label{eqn:overview}
\end{equation}

The non-repeated labels are obtained by the function $H$, as illustrated in Equation \ref{eq:non_overlapped_samples}, subsequently facilitating the extraction of the corresponding images.
\begin{equation}
        [Y_H] = [Y] \cap PD(k)
    \label{eq:non_overlapped_samples}
\end{equation}

Conversely, the second branch is inferred exclusively with the new class image $[X_H, Y_H]$ (i.e., non-repeated) samples and their corresponding exemplars $[X_{E\_H}, Y_{E\_H}]$, as shown in Equation \ref{eq:vm2}
\begin{equation}
    \begin{gathered}
        \hat{y}^{VM_2} = VM_2^{h_{i-1}} ([X_H,  X_{E\_H}]) \\
        prediction(x_i) = argmax(\hat{y}^{VM_2}_i) \\
        Accuracy^{VM_2} = \frac{1}{N_2}\sum_{i=1}^{N_2}[\hat{y}^{VM_2} = {y}^{VM_2}]
        \label{eq:vm2}
    \end{gathered}  
\end{equation}
where $\hat{y}^{VM_2} = [Y_H,  Y_{E\_H}]$ and  $N_2$ is the  corresponding batch size.

Following the initial task, the achieved accuracy is preserved as the threshold accuracy ($TH$), given that different food consumption patterns yield different accuracy and cannot be fixed. Subsequently, the DMU module employs this threshold accuracy to determine the branch with accuracy surpassing the threshold. The data from this branch is then directed into the main training pipeline. The primary model, trained until task $t_i$ and represented as $h_{i-1}$, is subsequently updated via supervised backpropagation.

% %%%%%%%%%%%%%%%%%%%%%%%%%%%%%%%%%%%%%%%%%%%%%%% %%%%%%%%%%%%%%%%%%%%%%%%%%%%%%%%%%%%%%%%%%%%%%
\section{Experiments}

%%%%%%%%%%%%%%%%%%%% table accuracy food101+vfn - exp
\begin{table*}
\centering
\caption{\textbf{Average Accuracy (\%)} rate with task sizes of 5, 10, and 20 with a 5K memory buffer on the Food-101 and VFN dataset with exponential distribution ($D$). Results include existing ER methods and the enhanced version using our method (DMU) leveraging the RDDM framework for realistic food image sequences. The best accuracy results are highlighted in \textbf{boldface}.}
\label{tab:acc_exp_food101_vfn}
\resizebox{\textwidth}{!}{
\begin{tabular}{l||ccc|ccc|ccc||ccc|ccc|ccc} 
\toprule
Dataset & \multicolumn{9}{c||}{Food-101} & \multicolumn{9}{c}{VFN} \\ 
\hline
Task size & \multicolumn{3}{c|}{5} & \multicolumn{3}{c|}{10} & \multicolumn{3}{c||}{20} & \multicolumn{3}{c|}{5} & \multicolumn{3}{c|}{10} & \multicolumn{3}{c}{20} \\ 
\hline
Category & Short & Mod & Long & Short & Mod & Long & Short & Mod & Long & \multicolumn{1}{c}{Short} & \multicolumn{1}{c}{Mod} & \multicolumn{1}{c|}{Long} & \multicolumn{1}{c}{Short} & \multicolumn{1}{c}{Mod} & \multicolumn{1}{c|}{Long} & \multicolumn{1}{c}{Short} & \multicolumn{1}{c}{Mod} & \multicolumn{1}{c}{Long} \\ 
\hline
 % & \multicolumn{18}{c}{} \\ 
\hline
Finetune & 53.00 & 51.00 & 48.65 & 45.23 & 37.50 & 35.10 & 22.00 & 23.67 & 21.11 & 33.10 & 25.50 & 26.90 & 23.67 & 20.80 & 19.17 & 14.62 & 11.84 & 10.80  \\ 
\hline
iCaRL [CVPR '17] & 75.02 & 69.34 & 57.75 & 67.0 & 56.6 & 49.65 & 57.24 & 51.69 & 47.13 & 59.04 & 52.01 & 38.39 & 50.78 & 46.64 & 36.07 & 44.09 & 47.96 & 34.81  \\
iCaRL + DMU & \textbf{76.04} & \textbf{70.68} & \textbf{59.22} & \textbf{67.14} & \textbf{62.29} & \textbf{52.19} & \textbf{59.72} & \textbf{53.65} & \textbf{50.37} & \textbf{59.77} & \textbf{53.36} & \textbf{42.95} & \textbf{52.72} & \textbf{46.93} & \textbf{36.17} & \textbf{46.66} & \textbf{48.31} & \textbf{34.88}  \\
\textbf{Gain $(\Delta)$} & 1.02 & 1.34 & 1.47 & 0.14 & 5.69 & 2.54 & 2.48 & 1.96 & 3.24 & 0.73 & 1.35 & 4.56 & 1.94 & 0.31 & 0.10 & 2.55 & 0.35 & 0.07 \\ 
\hline
MIR [NeurIPS '19] & 74.71 & 69.79 & 57.55 & 64.0 & 57.0 & 48.9 & 54.15 & 48.39 & 43.85 & 49.41 & 46.01 & 43.88 & 44.01 & 42.21 & 35.83 & 33.96 & 31.12 & 30.73 \\
MIR + DMU & \textbf{82.14} & \textbf{75.69} & \textbf{67.97} & \textbf{72.73} & \textbf{71.17} & \textbf{67.72} & \textbf{65.50} & \textbf{63.03} & \textbf{62.04} & \textbf{54.33} & \textbf{48.06} & \textbf{44.90} & \textbf{50.78} & \textbf{45.64} & \textbf{36.50} & \textbf{38.81} & \textbf{34.95} & \textbf{34.81}  \\
\textbf{Gain $(\Delta)$} & 7.43 & 5.9 & 10.42 & 8.73 & 14.17 & 18.82 & 11.35 & 14.64 & 18.19 & 4.92 & 1.96 & 1.02 & 6.77 & 3.43 & 0.67 & 4.92 & 3.83 & 4.08  \\ 
\hline
GSS [NeurIPS '19] & 72.17 & 69.77 & 58.54 & 51.5 & 50.5 & 47.8 & 40.33 & 42.8 & 38.51 & 53.41 & 45.64 & 37.19 & 44.68 & 44.13 & 34.49 & 33.74 & 34.75 & 29.82 \\
GSS + DMU~ ~ ~ & \textbf{76.29} & \textbf{74.01} & \textbf{68.19} & \textbf{55.07} & \textbf{66.17} & \textbf{66.06} & \textbf{47.43} & \textbf{49.86} & \textbf{50.13} & \textbf{55.59} & \textbf{47.52} & \textbf{38.83} & \textbf{48.10} & \textbf{47.91} & \textbf{38.35} & \textbf{39.44} & \textbf{35.70} & \textbf{30.08} \\
\textbf{Gain $(\Delta)$} & 4.12 & 4.24 & 9.65 & 3.57 & 15.67 & 18.26 & 4.10 & 7.06 & 11.62 & 2.18 & 1.88 & 1.64 & 3.42 & 3.78 & 3.86 & 5.70 & 0.96 & 0.26 \\ 
\hline
ASER [AAAI '21] & 77.87 & 72.51 & 62.89 & 67.70 & 66.6 & 61.12 & 59.70 & 58.29 & \textbf{56.35} & 51.86 & 48.56 & 42.19 & 43.34 & 42.32 & 36.01 & 33.68 & 32.14 & 28.61  \\
ASER + DMU & \textbf{78.33} & \textbf{73.96} & \textbf{64.06} & \textbf{72.27} & \textbf{68.17} & \textbf{64.06} & \textbf{60.59} & \textbf{59.74} & 56.084 & \textbf{52.78} & \textbf{51.15} & \textbf{46.08} & \textbf{43.59} & \textbf{44.47} & \textbf{40.36} & \textbf{33.74} & \textbf{32.75} & \textbf{30.07}  \\
\textbf{Gain $(\Delta)$} & 0.46 & 1.45 & 1.17 & 4.57 & 2.1 & 2.94 & 0.89 & 1.45 & – & 0.92 & 2.59 & 3.89 & 0.25 & 2.15 & 4.35 & 0.06 & 0.61 & 1.46 \\ 
\hline
DVC [CVPR '22] & 77.06 & 72.86 & 64.43 & 73.46 & 66.89 & 61.05 & 58.74 & 60.51 & 59.90 & 48.55 & 39.88 & 38.57 & 41.64 & 29.96 & 30.53 & 30.93 & 34.56 & 27.66 \\DVC + DMU & \textbf{78.53} & \textbf{74.96} & \textbf{65.95} & \textbf{74.52} & \textbf{68.75} & \textbf{62.15} & \textbf{60.51} & \textbf{61.82} & \textbf{59.94} & \textbf{50.06} & \textbf{40.63} & \textbf{41.28} & \textbf{44.23} & \textbf{39.59} & \textbf{32.87} & \textbf{36.81} & \textbf{35.26} & \textbf{30.32} \\
\textbf{Gain $(\Delta)$} & 1.47 & 1.35 & 1.53 & 1.06 & 1.86 & 1.1 & 1.77 & 1.82 & 0.04 & 1.51 & 0.75 & 2.71 & 2.59 & 9.63 & 2.34 & 5.88 & 0.7 & 2.66 \\
\hline
\bottomrule
\end{tabular}
}
\end{table*}

\subsection{Experimental setup}
\label{subsec:exp_setup}
\textbf{Datasets.} We introduce two benchmarks for Realistic Online Class Incremental Learning (R-OCIL) in the realm of real-world food image classification: the Food-101 \cite{food101} dataset (101 classes) and the VFN~\cite{he2022longtailed} dataset (74 classes). We simulate food consumption patterns by utilizing the RDDM module, which categorizes food classes into short-term, moderate-term, and long-term, as discussed in Section \ref{subsec:formulation}.
The Food-101 dataset comprises 101,000 images with 750 training images per class. The VFN dataset contains over 15,000 training images across 74 classes, representing commonly consumed food categories in the United States based on the WWEIA database~\footnote{https://data.nal.usda.gov/dataset/what-we-eat-america-wweia-database}. Test sets are balanced, with 250 images per class for Food-101 and 50 images per class for VFN. Our experiment involves assessing 5, 10, and 20 tasks using a 5K memory buffer, facilitating performance evaluation of all considered methods.

\textbf{Baselines.} Our analysis will delve into the performance of five prevalent ER OCIL methods, namely iCaRL~\cite{icarl}, MIR~\cite{MIR}, GSS~\cite{GSS}, ASER~\cite{ASER}, DVC~\cite{DVC}. These methods are discussed in Section~\ref{sec:related_work} and are used for evaluation within realistic food consumption scenarios simulated by RDDM. A comparison will be drawn between these methods and their enhanced versions, incorporating our proposed plug-and-play DMU. We incorporate a baseline method dubbed \textbf{Fine-tune} for reference. This method exclusively employs data from new classes and cross-entropy loss for continual learning without considering prior task performance. Hence, it serves as a performance lower bound. It is crucial to emphasize that our comparison does not encompass ``offline" methods centered on repetition.  This is due to the fundamental distinction between ``online" and ``offline" settings in terms of data encounter, training frameworks, and buffer strategies, as outlined in Section~\ref{sec:intro}.

\textbf{Evaluation Metrics.} For a fair evaluation, we consider access to a held-out test set for each of the $T$ tasks. Upon the model's completion of learning task $t_i$, we assess its test performance across all $T$ tasks. This process allows us to construct the matrix $B \in \mathbb{R}^{T \times T}$, where $B_{i,j}$ represents the test classification accuracy of the model for task $t_j$ after observing the final sample from task $t_i$~\cite{GEM}. Letting $\widehat{b}$ denote the vector of test accuracy for each task during random initialization, and we use the following metrics for performance evaluation: $Average\ Accuracy =  \frac{1}{T}\sum_{j=1}^{T} B_{T,j}$ where $B_{i, j}$ is the accuracy on task $j$ after the model has been trained from task $1$ through $i$.

\textbf{Implementation Details. }We use ResNet-18 \cite{resnet} pre-trained on ImageNet \cite{Krizhevsky} as the backbone structure for all experiments in comparison, and our implementation is based on PyTorch~\cite{pytorch}. Pre-trained networks simplify training and reduce training time, especially using complex datasets like Food-101 \cite{food101} and VFN~\cite{he2022longtailed}. The input image size for Food-101 and VFN is $224 \times 224$. For optimization, we use a stochastic gradient descent optimizer with a fixed learning rate of 0.001 and a weight decay of $10^{-4}$. The training batch size is 16, while the testing batch size is 128. We set the distribution $D$ (referring to Section \ref{subsec:rddm}) as exponential. Throughout the training process, the model encounters the incoming data, excluding exemplars, only once across all experimental iterations. In order to evaluate different experience replay methods, we establish a memory buffer size of 5K for storing exemplars during training for knowledge rehearsal. We conduct experiments by varying task sizes (5, 10, and 20) within all food consumption categories, as explained in Section~\ref{subsec:formulation}.

\subsection{Experimental Results}
\label{subsec:exp_res}

Table~\ref{tab:acc_exp_food101_vfn} presents the average accuracy across all task sizes and the three categories of food consumption patterns in both Food-101 and VFN datasets. It is evident that the continual learning performance varies considerably for different food consumption categories and task sizes. For instance, the model has fewer overall classes to learn in the short-term food consumption pattern with the fewest classes and the smallest task size. Consequently, even with fewer repetitions, the model achieves higher accuracy and less catastrophic forgetting. In contrast, we observe the lowest performance figures as the task sizes increase and accuracy is measured in the long-term food consumption category. This decrease in accuracy is because the model attempts to learn the most classes present across many tasks (as $\alpha$ reduces and $\beta$ increases), leading to lower accuracy and higher forgetting, despite the increase in repeated classes. Specifically, we observe severe catastrophic forgetting problems when using Finetune due to the lack of training data for learned classes during the continual learning process. All the existing ER methods outperform Finetune since they store a small sample of learned training classes for knowledge rehearsal during the continual learning process. However, when we enhance the existing ER methods with our proposed DMU module, we observe a substantial increase in performance consistently across all categories and task sizes in both Food-101 and VFN datasets. Our DMU module selects the ideal data for training by comparing the performance of the entire training batch with the new class data, leading to higher learning and lower catastrophic forgetting.

We immediately observe that the performance on the VFN dataset is notably lower than that of the Food-101 dataset, as seen in Table~\ref{tab:acc_exp_food101_vfn}. This difference in performance is due to the VFN dataset's smaller size, with only 15,000 training images across 74 categories, compared to the Food-101 dataset, which has 75,750 training images across 101 categories. Furthermore, the VFN dataset is an inherently imbalanced real-world dataset, which results in lower accuracy overall. Nevertheless, including the DMU module in the existing ER methodologies still significantly enhances performance across all experimented categories. On average, we demonstrate a 26\% increase on the Food-101 dataset and a 12.5\% increase on the VFN dataset, respectively.

\subsection{Ablation Study}
\label{subsec:ablation}
In this section, we perform ablation studies to assess the effectiveness of our DMU module across varying memory buffers (1K and 0.5K) and different data distributions $D$ simulated by RDDM (Section~\ref{subsec:rddm}, Equation~\ref{eqn:class_num}). Our focus for the data distribution is on a scenario exhibiting reduced imbalance. To achieve this, we configure $D$ as Gaussian with a mean of zero and a standard deviation of 0.4, aiming to simulate a broader spectrum with decreased imbalance effects. We apply the DVC~\cite{DVC} method, utilizing a consistent exemplar size of 5K and a task size of 10 for the Food-101 dataset.
To assess the impact of different buffer sizes on ER methods and the effectiveness of our proposed model across these sizes and present the results in Table~\ref{table:buffer_ablation}.

%Distribution ablation table
\begin{table}
\centering
\caption{\textbf{Average Accuracy (\%)} on Food-101 using DVC, and enhanced DVC with DMU by varying data distributions with a 5K buffer. The best results are marked in bold.}
\label{tab:acc_comp_exp_gauss}
\resizebox{0.65\columnwidth}{!}{
\begin{tabular}{c|c|c} 
\toprule
\textbf{Method} & \textbf{Distribution} & \textbf{Accuracy} \\ 
\midrule
DVC & Exp & 66.89 \\
DVC+DMU & Exp & \textbf{68.75} \\
Gain $(\Delta)$ &  & \textbf{1.86} \\ 
\midrule
DVC & Gauss & 60.89 \\
DVC+DMU & Gauss & \textbf{64.55} \\
Gain $(\Delta)$ &  & \textbf{3.75} \\
\bottomrule
\end{tabular}
}
\end{table}

%Imbalance ablation table
\begin{table}
\caption{\textbf{Average Accuracy (\%)} on VFN using best performing methods on varied buffer sizes. The best results are marked in bold.}
\label{table:buffer_ablation}
\resizebox{0.95\columnwidth}{!}{
\begin{tabular}{@{}l|cccccc@{}}
\toprule
Dataset     & \multicolumn{6}{c}{VFN}                                                                                                  \\ \hline
Buffer size & \multicolumn{3}{c|}{2K}                                               & \multicolumn{3}{c}{0.5K}                         \\ \hline
Task size   & \multicolumn{3}{c|}{20}                                               & \multicolumn{3}{c}{20}                           \\ \hline
Category    & short          & med            & \multicolumn{1}{c|}{long}           & short          & med            & long           \\ \hline
MIR         & 30.50          & 28.93          & \multicolumn{1}{c|}{28.22}          & 27.81          & 24.72          & 26.24          \\
MIR + DMU   & \textbf{31.17} & \textbf{30.31} & \multicolumn{1}{c|}{\textbf{29.63}} & \textbf{29.39} & \textbf{27.60} & \textbf{26.33} \\
Gain $(\Delta)$        & 0.67           & 1.38           & \multicolumn{1}{c|}{1.41}           & 1.58           & 2.88           & 0.09           \\ \hline
ASER        & 33.29          & 33.78          & \multicolumn{1}{c|}{29.31}          & 29.21          & 30.51          & 27.61          \\
ASER+DMU    & \textbf{33.86} & \textbf{34.70} & \multicolumn{1}{c|}{\textbf{29.48}} & \textbf{29.78} & \textbf{30.92} & \textbf{28.26} \\
Gain $(\Delta)$        & 0.57           & 0.92           & \multicolumn{1}{c|}{0.17}           & 0.57           & 0.41           & 0.65           \\ \hline
DVC         & 21.66          & 20.02          & \multicolumn{1}{c|}{19.00}          & 20.95          & 19.73          & 18.80          \\
DVC + DMU   & \textbf{22.70} & \textbf{21.22} & \multicolumn{1}{c|}{\textbf{18.42}} & \textbf{21.08} & \textbf{21.06} & \textbf{19.41} \\
Gain $(\Delta)$        & 1.04           & 1.20           & \multicolumn{1}{c|}{0.58}           & 0.13           & 1.33           & 0.61           \\ \hline
GSS         & 27.28          & 29.32          & \multicolumn{1}{c|}{29.20}          & 22.75          & 26.47          & 25.06          \\
GSS + DMU   & \textbf{28.20} & \textbf{32.29} & \multicolumn{1}{c|}{\textbf{29.22}} & \textbf{24.49} & \textbf{30.85} & \textbf{25.98} \\
Gain $(\Delta)$        & 0.92           & 2.97           & \multicolumn{1}{c|}{0.02}           & 1.74           & 4.38           & 0.92           \\ \hline
iCaRL       & 42.50          & 34.00          & \multicolumn{1}{c|}{33.60}          & 36.28          & 25.60          & 26.22          \\
iCaRL + DMU & \textbf{46.34} & \textbf{41.67} & \multicolumn{1}{c|}{\textbf{45.25}} & \textbf{37.74} & \textbf{36.67} & \textbf{39.04} \\
Gain $(\Delta)$        & 3.84           & 7.67           & \multicolumn{1}{c|}{11.65}          & 1.46           & 11.07          & 12.82          \\ \hline
\end{tabular}
}
\end{table}

\begin{figure}[t]
    \centering
    \includegraphics[width=.9\columnwidth]{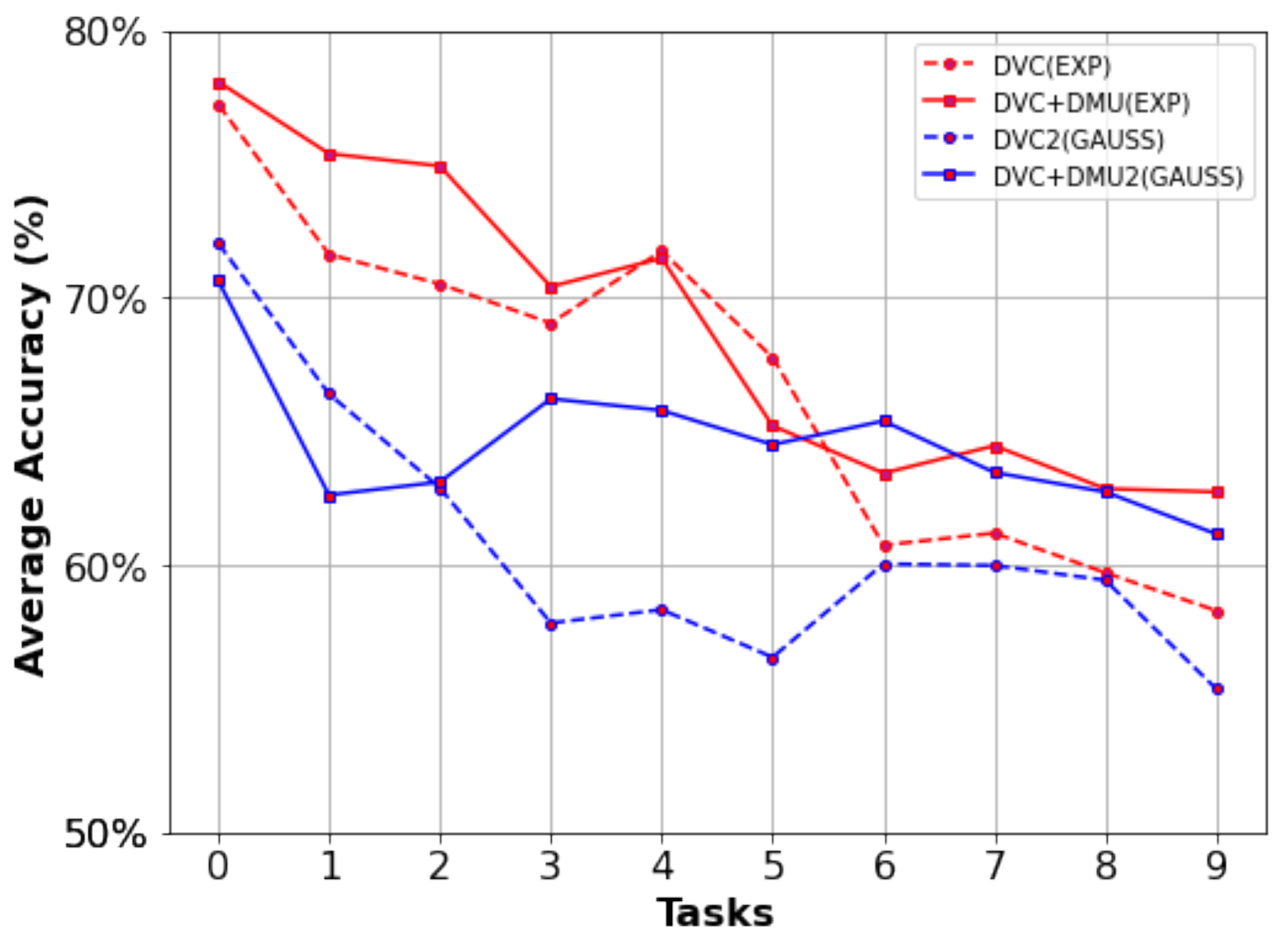}
    \caption{Accuracy of each incremental task in the Gaussian(GAUSS) and Exponential(EXP) distributions for the moderate food consumption category on DVC (with our proposed DMU module) using the Food-101 dataset with 10 tasks.}
    \label{fig:gauss_exp_comp}
\end{figure}

The results presented in Table~\ref{tab:acc_comp_exp_gauss} highlight the significant enhancement our DMU module brings to the performance of existing ER methods on the Food-101 dataset, even in the presence of varying levels of imbalance (as demonstrated in Figure~\ref{fig:gauss_exp_comp}). These improvements are consistent across different buffer sizes, as highlighted in Table~\ref{table:buffer_ablation}. Traditional ER techniques, designed with balanced tasks and uniform sample distributions, are notably affected by the impact of input stream repetitions and imbalances, resulting in performance deterioration. These findings emphasize the effectiveness of targeted training with optimal samples compared to training with the complete set of input samples, irrespective of the specific ER method employed.
% %%%%%%%%%%%%%%%%%%%%%%%%%%%%%%%%%%%%%%%%%%%%%%% %%%%%%%%%%%%%%%%%%%%%%%%%%%%%%%%%%%%%%%%%%%%%%
\section{Conclusion}
\label{sec:conclusion}

This paper focuses on Online Class-Incremental Learning for real-world food image classification. We introduce a novel Realistic Data Distribution Module (RDDM) to simulate real-world food consumption patterns in different scenarios along with a plug-and-play Dynamic Model Update (DMU) module, which is compatible with existing ER methods by independently targeting the training pipeline for improving performance. We show that our data distribution considers more realistic scenarios than existing OCIL systems, including a corresponding experimental benchmark. Furthermore, experimental results demonstrate our proposed method outperforms existing methods in more realistic food image classification settings on the challenging Food-101 and VFN datasets. Our future work will focus on improving the exemplar storage procedure by exploiting the most representative samples simulated by the RDDM framework. We also plan to evaluate our method on large supervised open-world datasets in the future.

%%%%%%%%% REFERENCES
{\small
\bibliographystyle{ieee_fullname}
\bibliography{main}

\begin{thebibliography}{10}\itemsep=-1pt

\bibitem{record-diet-assess}
Ziad Ahmad, Marc Bosch, Nitin Khanna, Deborah~A Kerr, Carol~J Boushey, Fengqing Zhu, and Edward~J Delp.
\newblock A mobile food record for integrated dietary assessment.
\newblock {\em Proceedings of the 2nd International Workshop on Multimedia Assisted Dietary Management}, page 41–49, October 2016.

\bibitem{MIR}
Rahaf Aljundi, Lucas Caccia, Eugene Belilovsky, Massimo Caccia, Min Lin, Laurent Charlin, and Tinne Tuytelaars.
\newblock {\em Online Continual Learning with Maximally Interfered Retrieval}.
\newblock Curran Associates Inc., 2019.

\bibitem{GSS}
Rahaf Aljundi, Min Lin, Baptiste Goujaud, and Yoshua Bengio.
\newblock Gradient based sample selection for online continual learning.
\newblock {\em Advances in Neural Information Processing Systems}, 32, 2019.

\bibitem{Belouadah2019IL2MCI}
Eden Belouadah and Adrian~Daniel Popescu.
\newblock Il2m: Class incremental learning with dual memory.
\newblock {\em 2019 IEEE/CVF International Conference on Computer Vision (ICCV)}, pages 583--592, 2019.

\bibitem{food101}
Lukas Bossard, Matthieu Guillaumin, and Luc Van~Gool.
\newblock Food-101 -- mining discriminative components with random forests.
\newblock {\em European Conference on Computer Vision}, 112(1):1--21, 2015.

\bibitem{cil1}
Francisco~M. Castro, Manuel~J. Mar{\'i}n-Jim{\'e}nez, Nicol{\'a}s Guil, Cordelia Schmid, and Karteek Alahari.
\newblock End-to-end incremental learning.
\newblock {\em Proceedings of the European Conference on Computer Vision}, pages 241--257, 2018.

\bibitem{intransigence}
Arslan Chaudhry, Puneet~K. Dokania, Thalaiyasingam Ajanthan, and Philip H.~S. Torr.
\newblock Riemannian walk for incremental learning: Understanding forgetting and intransigence.
\newblock {\em Proceedings of the European Conference on Computer Vision (ECCV)}, September 2018.

\bibitem{AGEM}
Arslan Chaudhry, Marc’Aurelio Ranzato, Marcus Rohrbach, and Mohamed Elhoseiny.
\newblock Efficient lifelong learning with a-{GEM}.
\newblock {\em International Conference on Learning Representations}, 2019.

\bibitem{pmlr-v119-chrysakis20a}
Aristotelis Chrysakis and Marie-Francine Moens.
\newblock Online continual learning from imbalanced data.
\newblock {\em Proceedings of the 37th International Conference on Machine Learning}, 119:1952--1961, 13--18 Jul 2020.

\bibitem{cil_enough}
Andrea Cossu, Gabriele Graffieti, Lorenzo Pellegrini, Davide Maltoni, Davide Bacciu, Antonio Carta, and Vincenzo Lomonaco.
\newblock Is class-incremental enough for continual learning?
\newblock {\em Frontiers in Artificial Intelligence}, 5, 2022.

\bibitem{diet_1}
Denise de Ridder, Floor Kroese, Catharine Evers, Marieke Adriaanse, and Marleen Gillebaart.
\newblock Healthy diet: Health impact, prevalence, correlates, and interventions.
\newblock {\em Psychol. Health}, 32(8):907--941, Aug. 2017.

\bibitem{dietary-pattern-assess}
Heather~A Eicher-Miller, Saul Gelfand, Youngha Hwang, Edward Delp, Anindya Bhadra, and Jiaqi Guo.
\newblock Distance metrics optimized for clustering temporal dietary patterning among {U.S}. adults.
\newblock {\em Appetite}, 144(104451):104451, Jan. 2020.

\bibitem{cat_forget}
Ian Goodfellow, Mehdi Mirza, Xia Da, Aaron Courville, and Y. Bengio.
\newblock An empirical investigation of catastrophic forgeting in gradient-based neural networks.
\newblock 12 2013.

\bibitem{DVC}
Yanan Gu, Xu Yang, Kun Wei, and Cheng Deng.
\newblock Not just selection, but exploration: Online class-incremental continual learning via dual view consistency.
\newblock {\em Proceedings of the IEEE/CVF Conference on Computer Vision and Pattern Recognition (CVPR)}, pages 7442--7451, June 2022.

\bibitem{yue_food_real_world}
Yue Han, Sri Yarlagadda, Tonmoy Ghosh, Fengqing Zhu, Edward Sazonov, and Edward Delp.
\newblock Improving food detection for images from a wearable egocentric camera.
\newblock 01 2023.

\bibitem{food_img_cls_deep_dnn}
Hamid Hassannejad, Guido Matrella, Paolo Ciampolini, Ilaria De~Munari, Monica Mordonini, and Stefano Cagnoni.
\newblock Food image recognition using very deep convolutional networks.
\newblock In {\em Proceedings of the 2nd International Workshop on Multimedia Assisted Dietary Management}, MADiMa '16, page 41–49, New York, NY, USA, 2016. Association for Computing Machinery.

\bibitem{ocil_orig_1}
Tyler Hayes, Nathan Cahill, and Christopher Kanan.
\newblock Memory efficient experience replay for streaming learning.
\newblock {\em IEEE Robotics and Automation Letters}, 4(10):9769--9776, 2019.

\bibitem{he2022longtailed}
Jiangpeng He, Luotao Lin, Heather Eicher-Miller, and Fengqing Zhu.
\newblock Long-tailed food classification, 2022.

\bibitem{he2023long}
Jiangpeng He, Luotao Lin, Jack Ma, Heather~A Eicher-Miller, and Fengqing Zhu.
\newblock Long-tailed continual learning for visual food recognition.
\newblock {\em arXiv preprint arXiv:2307.00183}, 2023.

\bibitem{CIL_online}
Jiangpeng He, Runyu Mao, Zeman Shao, and Fengqing Zhu.
\newblock Incremental learning in online scenario.
\newblock {\em Proceedings of the IEEE/CVF Conference on Computer Vision and Pattern Recognition}, pages 13923--13932, 06 2020.

\bibitem{food_ocil}
Jiangpeng He and Fengqing Zhu.
\newblock Online continual learning for visual food classification.
\newblock {\em Proceedings of the IEEE/CVF International Conference on Computer Vision Workshops}, pages 2337--2346, October 2021.

\bibitem{resnet}
Kaiming He, Xiangyu Zhang, Shaoqing Ren, and Jian Sun.
\newblock Deep residual learning for image recognition.
\newblock {\em Proceedings of the IEEE/CVF Conference on Computer Vision and Pattern Recognition}, pages 770--778, 2016.

\bibitem{hemati2023classincremental}
Hamed Hemati, Andrea Cossu, Antonio Carta, Julio Hurtado, Lorenzo Pellegrini, Davide Bacciu, Vincenzo Lomonaco, and Damian Borth.
\newblock Class-incremental learning with repetition, 2023.

\bibitem{cil2}
Saihui Hou, Xinyu Pan, Chen~Change Loy, Zilei Wang, and Dahua Lin.
\newblock Learning a unified classifier incrementally via rebalancing.
\newblock {\em Proceedings of the IEEE/CVF Conference on Computer Vision and Pattern Recognition}, June 2019.

\bibitem{EWC}
James Kirkpatrick, Razvan Pascanu, Neil Rabinowitz, Joel Veness, Guillaume Desjardins, Andrei~A. Rusu, Kieran Milan, John Quan, Tiago Ramalho, Agnieszka Grabska-Barwinska, Demis Hassabis, Claudia Clopath, Dharshan Kumaran, and Raia Hadsell.
\newblock Overcoming catastrophic forgetting in neural networks.
\newblock {\em Proceedings of the National Academy of Sciences}, 114(13):3521--3526, 2017.

\bibitem{diet_approach_durations}
Chrysi Koliaki, Theodoros Spinos, $M$arianna Spinou, $M$aria-Eugenia Brinia, Dimitra Mitsopoulou, and Nicholas Katsilambros.
\newblock Defining the optimal dietary approach for safe, effective and sustainable weight loss in overweight and obese adults.
\newblock {\em Healthcare (Basel)}, 6(3):73, June 2018.

\bibitem{Krizhevsky}
Alex Krizhevsky, Ilya Sutskever, and Geoffrey~E. Hinton.
\newblock Imagenet classification with deep convolutional neural networks.
\newblock {\em ACM}, 60(6):84–90, may 2017.

\bibitem{LWF}
Zhizhong Li and Derek Hoiem.
\newblock Learning without forgetting.
\newblock {\em IEEE Transactions on Pattern Analysis and Machine Intelligence}, 40(12):2935--2947, 2018.

\bibitem{PCR}
Huiwei Lin, Baoquan Zhang, Shanshan Feng, Xutao Li, and Yunming Ye.
\newblock Pcr: Proxy-based contrastive replay for online class-incremental continual learning.
\newblock {\em Proceedings of the IEEE/CVF Conference on Computer Vision and Pattern Recognition}, 2023.

\bibitem{health-monitor}
L. Lin, J. Guo, M.~M. Aqeel, S.~B. Gelfand, E.~J. Delp, A. Bhadra, E.~A. Richards, E. Hennessy, and H.~A. Eicher-Miller.
\newblock {{J}oint temporal dietary and physical activity patterns: associations with health status indicators and chronic diseases}.
\newblock {\em Am J Clin Nutr}, 115(2):456--470, 02 2022.

\bibitem{Liu2016DeepFoodDL}
Chang Liu, Yu Cao, Yan Luo, Guanling Chen, Vinod Vokkarane, and Yunsheng Ma.
\newblock Deepfood: Deep learning-based food image recognition for computer-aided dietary assessment.
\newblock In {\em International Conference on Smart Homes and Health Telematics}, 2016.

\bibitem{better_weight_cons}
Xialei Liu, Marc Masana, Luis Herranz, Joost Van~de Weijer, Antonio~M. López, and Andrew~D. Bagdanov.
\newblock Rotate your networks: Better weight consolidation and less catastrophic forgetting.
\newblock {\em Proceedings of the International Conference on Pattern Recognition (ICPR)}, pages 2262--2268, 2018.

\bibitem{GEM}
David Lopez-Paz and Marc'Aurelio Ranzato.
\newblock Gradient episodic memory for continual learning.
\newblock {\em Proceedings of the 31st International Conference on Neural Information Processing Systems}, pages 6470--6479, 2017.

\bibitem{ocil_survey}
Zheda Mai, Ruiwen Li, Jihwan Jeong, David Quispe, Hyunwoo Kim, and Scott Sanner.
\newblock Online continual learning in image classification: An empirical survey.
\newblock {\em Neurocomput.}, 469(C):28–51, jan 2022.

\bibitem{mao2021_nutri_hierarchy}
Runyu Mao, Jiangpeng He, Luotao Lin, Zeman Shao, Heather~A. Eicher-Miller, and Fengqing Zhu.
\newblock Improving dietary assessment via integrated hierarchy food classification.
\newblock {\em 2021 IEEE 23rd International Workshop on Multimedia Signal Processing}, pages 1--6, 2021.

\bibitem{food_recog}
Runyu Mao, Jiangpeng He, Zeman Shao, Sri~Kalyan Yarlagadda, and Fengqing Zhu.
\newblock Visual aware hierarchy based food recognition.
\newblock {\em Proceedings of the Pattern Recognition/ ICPR International Workshops and Challenges}, pages 571--598, 2021.

\bibitem{dl_food_class4}
Runyu Mao, Jiangpeng He, Zeman Shao, Sri~Kalyan Yarlagadda, and Fengqing Zhu.
\newblock Visual aware hierarchy based food recognition.
\newblock {\em Proceedings of the Pattern Recognition. ICPR International Workshops and Challenges}, 12814:571–598, 2021.

\bibitem{CIL_survey}
M. Masana, X. Liu, B. Twardowski, M. Menta, A.~D. Bagdanov, and J. van~de Weijer.
\newblock Class-incremental learning: Survey and performance evaluation on image classification.
\newblock {\em IEEE Transactions on Pattern Analysis and; Machine Intelligence}, 45(05):5513--5533, may 2023.

\bibitem{catastrophic_forget_org}
Michael McCloskey and Neal~J. Cohen.
\newblock Catastrophic interference in connectionist networks: The sequential learning problem.
\newblock volume~24 of {\em Psychology of Learning and Motivation}, pages 109--165. Academic Press, 1989.

\bibitem{auto_diet_assess}
Simon Mezgec, Tome Eftimov, Tamara Bucher, and Barbara Koroušić~Seljak.
\newblock Mixed deep learning and natural language processing method for fake-food image recognition and standardization to help automated dietary assessment.
\newblock {\em Public Health Nutrition}, 22(7):1193–1202, 2019.

\bibitem{NutriNet}
Simon Mezgec and Barbara Koroušić~Seljak.
\newblock Nutrinet: A deep learning food and drink image recognition system for dietary assessment.
\newblock {\em Nutrients}, 9(7), 2017.

\bibitem{gcil}
Fei Mi, Lingjing Kong, Tao Lin, Kaicheng Yu, and Boi Faltings.
\newblock Generalized class incremental learning.
\newblock {\em 2020 IEEE/CVF Conference on Computer Vision and Pattern Recognition Workshops}, pages 970--974, 2020.

\bibitem{pytorch}
Adam Paszke, Sam Gross, Francisco Massa, Adam Lerer, James Bradbury, Gregory Chanan, Trevor Killeen, Zeming Lin, Natalia Gimelshein, Luca Antiga, Alban Desmaison, Andreas Kopf, Edward Yang, Zachary DeVito, Martin Raison, Alykhan Tejani, Sasank Chilamkurthy, Benoit Steiner, Lu Fang, Junjie Bai, and Soumith Chintala.
\newblock Pytorch: An imperative style, high-performance deep learning library.
\newblock pages 8024--8035. Curran Associates, Inc., 2019.

\bibitem{nutrition_dynamics}
Barry~M Popkin.
\newblock Global nutrition dynamics: the world is shifting rapidly toward a diet linked with noncommunicable diseases2.
\newblock {\em The American Journal of Clinical Nutrition}, 84(2):289--298, 2006.

\bibitem{icarl}
Sylvestre-Alvise Rebuffi, Alexander Kolesnikov, Georg Sperl, and Christoph~H. Lampert.
\newblock icarl: Incremental classifier and representation learning.
\newblock {\em Proceedings of the IEEE/CVF Conference on Computer Vision and Pattern Recognition}, pages 5533--5542, 2017.

\bibitem{foodai_logging}
Doyen Sahoo, Wang Hao, Shu Ke, Wu Xiongwei, Hung Le, Palakorn Achananuparp, Ee-Peng Lim, and Steven C.~H. Hoi.
\newblock Foodai: Food image recognition via deep learning for smart food logging.
\newblock In {\em Proceedings of the 25th ACM SIGKDD International Conference on Knowledge Discovery and Data Mining}, KDD '19, page 2260–2268, New York, NY, USA, 2019. Association for Computing Machinery.

\bibitem{dl_food_class2}
Zeman Shao, Shaobo Fang, Runyu Mao, Jiangpeng He, Janine Wright, Deborah Kerr, Carol~Jo Boushey, and Fengqing Zhu.
\newblock Towards learning food portion from monocular images with cross-domain feature adaptation, 2021.

\bibitem{ASER}
Dongsub Shim, Zheda Mai, Jihwan Jeong, Scott Sanner, Hyunwoo Kim, and Jongseong Jang.
\newblock Online class-incremental continual learning with adversarial shapley value.
\newblock {\em Proceedings of the AAAI Conference on Artificial Intelligence}, 35(11):9630--9638, 2021.

\bibitem{modern_trad_eating}
Gudrun Sproesser, Matthew~B Ruby, Naomi Arbit, Charity~S Akotia, Marle dos~Santos Alvarenga, Rachana Bhangaokar, Isato Furumitsu, Xiaomeng Hu, Sumio Imada, G{\"u}lbanu Kaptan, Martha Kaufer-Horwitz, Usha Menon, Claude Fischler, Paul Rozin, Harald~T Schupp, and Britta Renner.
\newblock Understanding traditional and modern eating: the {TEP10} framework.
\newblock {\em BMC Public Health}, 19(1):1606, Dec. 2019.

\bibitem{van_types_of_CL}
Gido~M van~de Ven, Tinne Tuytelaars, and Andreas~S Tolias.
\newblock Three types of incremental learning.
\newblock {\em Nat. Mach. Intell.}, 4(12):1185--1197, Dec. 2022.

\bibitem{dl_food_class3}
Hui Wu, Michele Merler, Rosario Uceda-Sosa, and John~R. Smith.
\newblock Learning to make better mistakes: Semantics-aware visual food recognition.
\newblock {\em Proceedings of the 24th ACM International Conference on Multimedia}, 2016:172--176, 2016.

\bibitem{synaptic_CL}
Friedemann Zenke, Ben Poole, and Surya Ganguli.
\newblock Continual learning through synaptic intelligence.
\newblock {\em Proceedings of the 34th International Conference on Machine Learning - Volume 70}, page 3987–3995, 2017.

\end{thebibliography}
}

\end{document}